\documentclass[times, twoside]{henriques-style}
\usepackage{blindtext}
\usepackage[nolist]{acronym}
\setcitestyle{square}
\usepackage{tikz}
\usetikzlibrary{positioning, calc, matrix, decorations.pathreplacing, shapes.geometric}
\usepackage{adjustbox}
\usepackage{url}

\usepackage{xspace}
\makeatletter
\DeclareRobustCommand\onedot{\futurelet\@let@token\@onedot}
\def\@onedot{\ifx\@let@token.\else.\null\fi\xspace}

\leadauthor{Salar} 

\begin{document}

\title{Cascade R-CNN for MIDOG Challenge}
\shorttitle{Mitotic Cell Detection Using Cascade R-CNN}

% Use letters for affiliations, numbers to show equal authorship (if applicable) and to indicate the corresponding author
\author[1]{Salar Razavi}
\author[1]{Fariba Dambandkhameneh}
\author[1]{Dimitri Androutsos}
\author[2,3]{Susan Done}
\author[1]{April Khademi}

\affil[1]{Image Analysis in Medicine Lab (IAMLAB), Electrical, Computer and Biomedical Engineering, Ryerson University, Toronto, ON, Canada}
\affil[2]{Princess Margaret Cancer Centre, University Health Network, Toronto, ON, Canada}
\affil[3]{Department of Laboratory Medicine and Pathobiology, University of Toronto, Toronto, ON, Canada}

\maketitle

%TC:break Abstract
%the command above serves to have a word count for the abstract
\begin{abstract}
Mitotic counts are one of the key indicators of breast cancer prognosis. However, accurate mitotic cell counting is still a difficult problem and is labourious. Automated methods have been proposed for this task, but are usually dependent on the training images and show poor performance on unseen domains. In this work, we present a multi-stage mitosis detection method based on a Cascade R-CNN developed to be sequentially more selective against false positives. On the preliminary test set, the algorithm scores an F$_1$~score of 0.7492.     
\end {abstract}
%TC:break main
%the command above serves to have a word count for the abstract

%\begin{keywords}
%Domain Generalization | Mitotic Count | Histopathology
%\end{keywords}

\begin{corrauthor}
salar.razavi@ryerson.ca
\end{corrauthor}

\section*{Introduction}
Breast cancer disease is a global concern affecting over 2 million women worldwide \cite{cancerReport}. Central to breast cancer diagnosis and treatment planning, is pathological analysis of tissue sections under magnification. There are typically three features that are used in grading, which includes mitosis counts, tubule formation and nuclear pleomorphism. Mitotic counts are a key indicator of tumour aggressiveness, but manual counting of mitosis is labourious, subjective and error prone.  With the advent of whole slide imaging (WSI) scanners, there is an opportunity to leverage computational algorithms to perform mitosis detection in an automatic and objective manner. However, a challenge with automated mitosis detection, and many computational pathology algorithms is domain shift \cite{stacke2019closer}. Different scanners and staining creates variability in colours and noise distributions, which can cause generalization challenges especially for deep learning-based algorithms when new data is out of the training distribution. To address these challenges, the \ac{midog} competition was launched \cite{marc_aubreville_2021_4573978} to test mitosis detection algorithms on images acquired from different scanners and laboratories.   In addition to the generalization challenges due to domain shift, discriminating mitotic figures from hard negative samples is also a big concern in mitosis detection. The proposed Cascade R-CNN based architecture is trained with images from different domains to detect mitotic figures and hard negative samples with high accuracy.

\section*{Material and Methods}
In this work, the proposed mitosis detection algorithm was developed using the official training set of the \ac{midog} dataset. The algorithm is based on a publicly available implementation of the Cascade R-CNN~\cite{cai2017cascade} which consists of a sequence of sequential detectors with increasing intersection over union (IoU) to reduce false positives which may be attributed to the hard to  detect mitotic cells. Because of small amount of images, progressively resampling in each stage is also used to reduce overfitting by ensuring there is a positive set of examples in each stage. These methods and datasets are detailed next. 
\subsection*{Dataset}
In total, the \ac{midog} dataset contains 150 annotated high power fields (HPFs). We extracted 18,960 patches of 512$\times$512 size from three scanners (Hamamatsu XR NanoZoomer 2.0, Hamamatsu S360, Aperio ScanScope CS2). Annotations consisted of two labels: one for the mitotic figures, and a second label for the hard-negative examples, which are darkly stained cells or regions that have similar appearance to mitosis (but are not mitosis). Only patches that had mitosis or hard-negative samples were used to train the models. In total, there were 3,072 training and 1,913 validation patches of 512$\times$512 size, respectively. The data was split randomly. In this subset of data, there were 2,437 mitotic figures and 1,558 hard negative examples. Considering the hard negative examples as an another class ensured the model to reduce the number of false positives specifically in dark-stained and low contrast images. 

\subsection*{Cascade R-CNN} In this work, the Cascade R-CNN architecture~\cite{cai2017cascade} is proposed for mitosis detection. The Cascade R-CNN model is a two-stage model that detects candidate regions (region proposal network), and a second stage that performs classification on the candidate regions (RPN+classification). The multi-scale nature of the Cascade R-CNN enables the detection of multiresolution structures by training with increasing IoU thresholds, which may  be more robust against false positives. Progressively sampling stage by stage improves detection and ensures that all detectors have a positive set of examples of equivalent size, and as a result reduces overfitting. Applying the same multi-stage procedure in the testing phase, enables a higher agreement between the hypotheses and the detector in each stage.

\subsection*{Network Training}
\begin{figure}[!ht]
\centering
\includegraphics[width=0.95\linewidth]{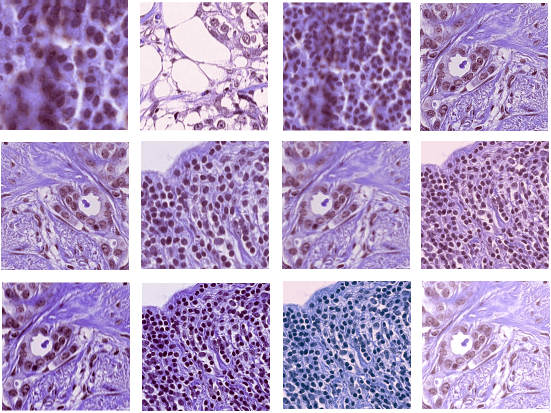}
\caption{Sample augmented training images.}
\label{fig:sampleimage}
\end{figure}

As previously described, 3,072 patches of size 512$\times$512 pixels with a batch size of 4 were used for training. All images were normalized using the Macenko stain normalization algorithm \cite{macenko2009method}. Data augmentation with random flipping, scaling, color, cropping and contrast was also considered. Through experimentation, it was found that random cropping (with minimum of 0.3 IoU of cropped patches) and scaling (to 4 other scales with $\pm$64 pixels steps) was optimal (See Fig. \ref{fig:sampleimage}). The two-stage Cascade R-CNN with ResNext101\_64x4d \cite{Xie2016} backbone pretrained on ImageNet \cite{5206848} dataset model with stochastic gradient descent (SGD) and a learning rate of 0.01 for 50 epochs. A linear warm up ratio of 0.001 for 500 steps was also applied to make training more stable. Gradient clipping was considered to prevent exploding gradients. 
To optimize training, different sampling methods were also considered. The IoU balanced \cite{wu2020ioubalanced} and online hard example mining (OHEM) \cite{shrivastava2016training} sampling methods were implemented to select hard samples according to their confidence. However, in comparison with random sampling performance worsened. This may be due to the similarities between mitotic figures (with high score) with some of the hard-negative annotations; or it could be related to the structure of the proposed two-stage detection method.  
As the Cascade R-CNN model is a two-stage detection (RPN+classification) architecture, added a focal loss \cite{lin2018focal} to overcome class imbalance but results did not improve and therefore, was not used. 
positive sample. As the model processes overlapping tiles, there may be multiple detections for a single mitosis. To overcome this, the output predictions were post-processed with non-maximum suppression \cite{neubeck2006efficient} and a 0.5 threshold is used to remove multiple overlapped bounding box detections. 
All the models are implemented with MMDetection \cite{mmdetection} library for automated detection on a RTX 2080 GTI GPU.

\section*{Evaluation and Results}
\begin{figure}[!ht]
\centering
\includegraphics[width=0.99\linewidth]{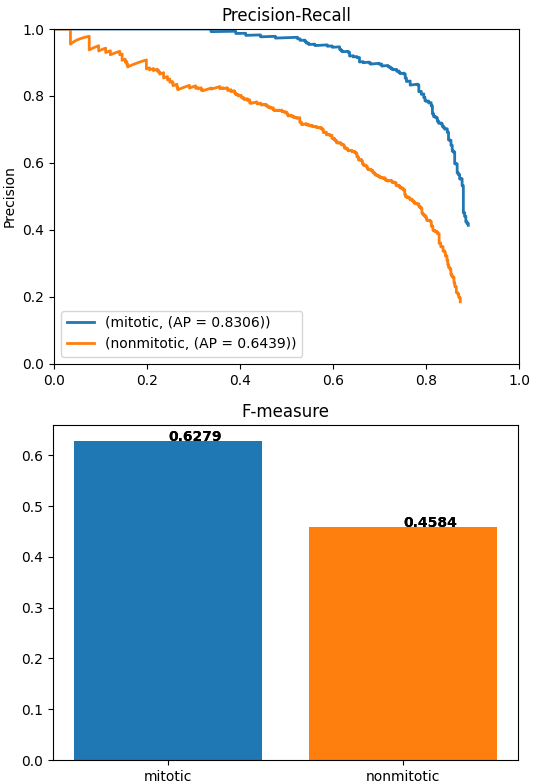}
\caption{Validation \acl{aucpr} per class for validation set.}
\label{fig:PR_Curve_each_class}
\end{figure}

The precision-recall curves for the validation set are shown in Fig. \ref{fig:PR_Curve_each_class} with an average precision (AP) of 0.8306 for the mitosis class and an AP of 0.6439 for the hard negative class.
The average F$_1$~score on the validation set was 0.63 and 0.46 for the mitosis and hard-negative samples, respectively.  Evaluation on the preliminary test set from the MIDOG organizers resulted in a mean F$_1$~score of 0.7492 (with 0.7707 precision and 0.7289 recall). 
\begin{figure}[!ht]
\centering
\includegraphics[width=0.99\linewidth]{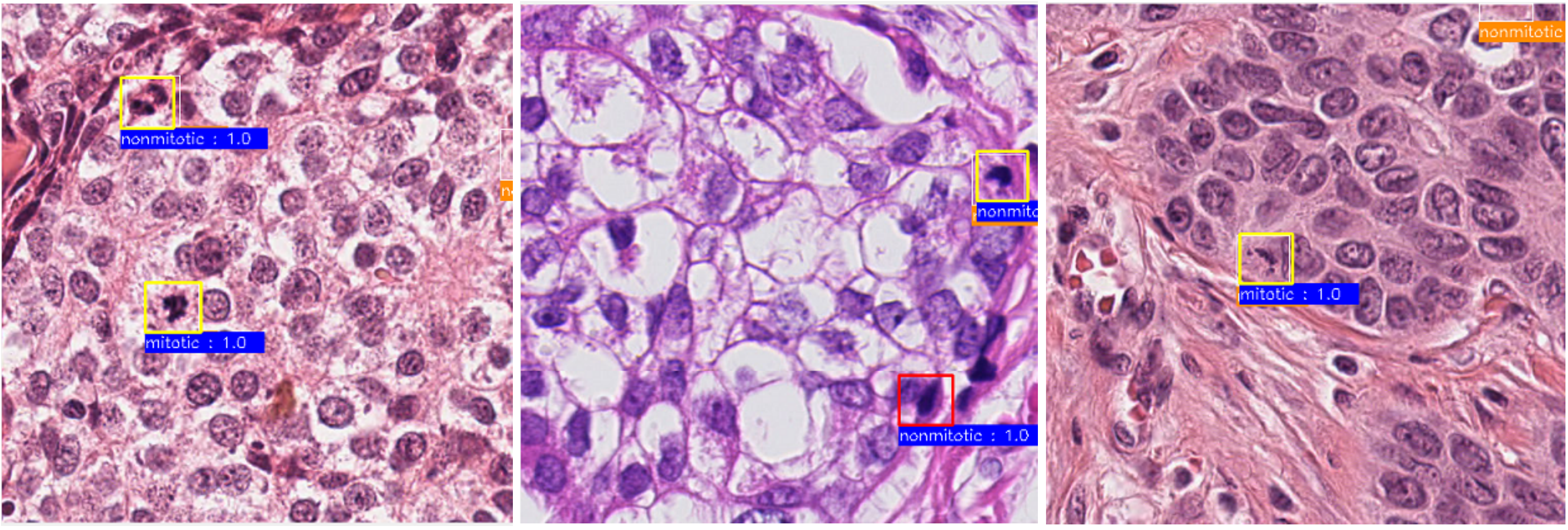}
\caption{Results of the detection module at patch-level. Yellow boxes highlights the true predictions, whereas red box shows the false predictions and white boxes are ground-truth annotations.
}
\label{fig:detectionresult}
\end{figure}
\section*{Discussion and Conclusion}
In this work, we presented an algorithm for the \ac{midog} challenge with a F$_1$~score of 0.6279 on the validation set and an F$_1$~score of 0.7492 on the preliminary test images. The model's performance on all of the preliminary test images are in the top but only for 003.tiff there are lots of false positives which degraded the overall performance.

\begin{acronym}
\acro{mc}[MC] {Mitotic Count}
\acro{midog}[MIDOG]{MItosis DOmain Generalization}
\acro{miccai}[MICCAI]{Medical Image Computing and Computer Assisted Intervention}
\acro{wsi}[WSI]{Whole Slide Image}
\acro{he}[H\&E]{Hematoxylin \& Eosin}
\acro{grl}[GRL]{Gradient Reverse Layer}
\acro{aucpr}[AUCPR]{area under the precision-recall curve}
\end{acronym}

\end{document}